\documentclass[journal]{IEEEtran}

\def\BibTeX{{\rm B\kern-.05em{\sc i\kern-.025em b}\kern-.08em
    T\kern-.1667em\lower.7ex\hbox{E}\kern-.125emX}}
\usepackage{textcomp}
\usepackage{graphicx}
\usepackage{cite}
\usepackage{amssymb}
\usepackage{hyperref}
\usepackage{multirow}
\usepackage{array}
\usepackage[acronym,toc,shortcuts]{glossaries}
\usepackage{url}
\usepackage{csquotes}
\usepackage{amsmath}
\usepackage{float}
\usepackage[acronym,toc,shortcuts]{glossaries}
\usepackage{color}
\usepackage{footmisc}
\usepackage{soul}

\usepackage{enumitem}

\usepackage[framemethod=tikz]{mdframed}
\usepackage[ruled,vlined,noend]{algorithm2e}

\begin{document}
%
\title{A Deep Neural Network for\\ SSVEP-based Brain-Computer Interfaces}
%
%
%

\author{Osman Berke Guney, Muhtasham Oblokulov, Huseyin Ozkan,\IEEEmembership{~Member,~IEEE} 
\thanks{O. B. Guney and H. Ozkan are with the Faculty of Engineering and Natural Sciences at Sabanci University, Istanbul, Turkey. Email: \{osmanberke, huseyin.ozkan\}@sabanciuniv.edu. M. Oblokulov is with the Technical University of Munich, Munich, Germany. Email: muhtasham.oblokulov@tum.de. This work was supported by The Scientific and Technological Research Council (TUBITAK) of Turkey under Contract 118E268. M. Oblokulov contributed to this study, when he was a senior student, as a part of his graduation project under the supervision of H. Ozkan.

Copyright (c) 2021 IEEE. Personal use of this material is permitted. However, permission to use this material for any other purposes must be obtained from the IEEE by sending an email to pubs-permissions@ieee.org.

Digital Object Identifier 10.1109/TBME.2021.3110440}}%

\maketitle
\begin{abstract}
\textit{Objective:} Target identification in brain-computer interface (BCI) spellers refers to the electroencephalogram (EEG) classification for predicting the target character that the subject intends to spell. When the visual stimulus of each character is tagged with a distinct frequency, the EEG records steady-state visually evoked potentials (SSVEP) whose spectrum is dominated by the harmonics of the target frequency. In this setting, we address the target identification and propose a novel deep neural network (DNN) architecture. \textit{Method:} The proposed DNN processes the multi-channel SSVEP with convolutions across the sub-bands of harmonics, channels, time, and classifies at the fully connected layer. We test with two publicly available large scale (\textit{the benchmark and BETA}) datasets consisting of in total $105$ subjects with $40$ characters. Our first stage training learns a global model by exploiting the statistical commonalities among all subjects, and the second stage fine tunes to each subject separately by exploiting the individualities. \textit{Results:} Our DNN achieves impressive information transfer rates (ITRs) on both datasets, \textit{265.23 bits/min} and \textit{196.59 bits/min}, respectively, with only $0.4$ seconds of stimulation. The code is available for reproducibility at \url{https://github.com/osmanberke/Deep-SSVEP-BCI}. \textit{Conclusion:} The presented DNN strongly outperforms the state-of-the-art techniques as our accuracy and ITR rates are the highest ever reported performance results on these datasets. \textit{Significance:} Due to its unprecedentedly high speller ITRs and flawless applicability to general SSVEP systems, our technique has great potential in various biomedical engineering settings of BCIs such as communication, rehabilitation and control.
\end{abstract}


%
\IEEEpeerreviewmaketitle

\section{Introduction}
%
%
%
%
\IEEEPARstart{B}{rain}-computer interfaces (BCIs) set up a direct communication channel between the human brain and a computer to translate brain signals to external commands \cite{tbme1}. BCIs have received increasing attention in recent years \cite{tbme2} and led to substantial progress  in many applications such as gaming \cite{gaming-bci}, stroke rehabilitation \cite{stroke_tbme3},   and cursor control \cite{cursor}. Another prominent application is the BCI speller \cite{P300_tpami} that assists patients with severe motor disabilities (e.g. amyotrophic lateral sclerosis), so that they can communicate by spelling via solely brain signals without muscular activity. BCI speller research is recently more focused on the use of steady-state visually evoked potentials (SSVEP) in EEG (electroencephalogram) signals \cite{trca,TS-CORRCA}
as the SSVEP has relatively higher signal-to-noise ratio (SNR) \cite{eegsnr}. Consequently, BCI SSVEP spellers achieve higher information transfer rate (ITR) with ease of system configuration \cite{ssvep-p300,cca-jfpm}.

\begin{figure*}[t!]
\centering
\includegraphics[width=0.90\textwidth]{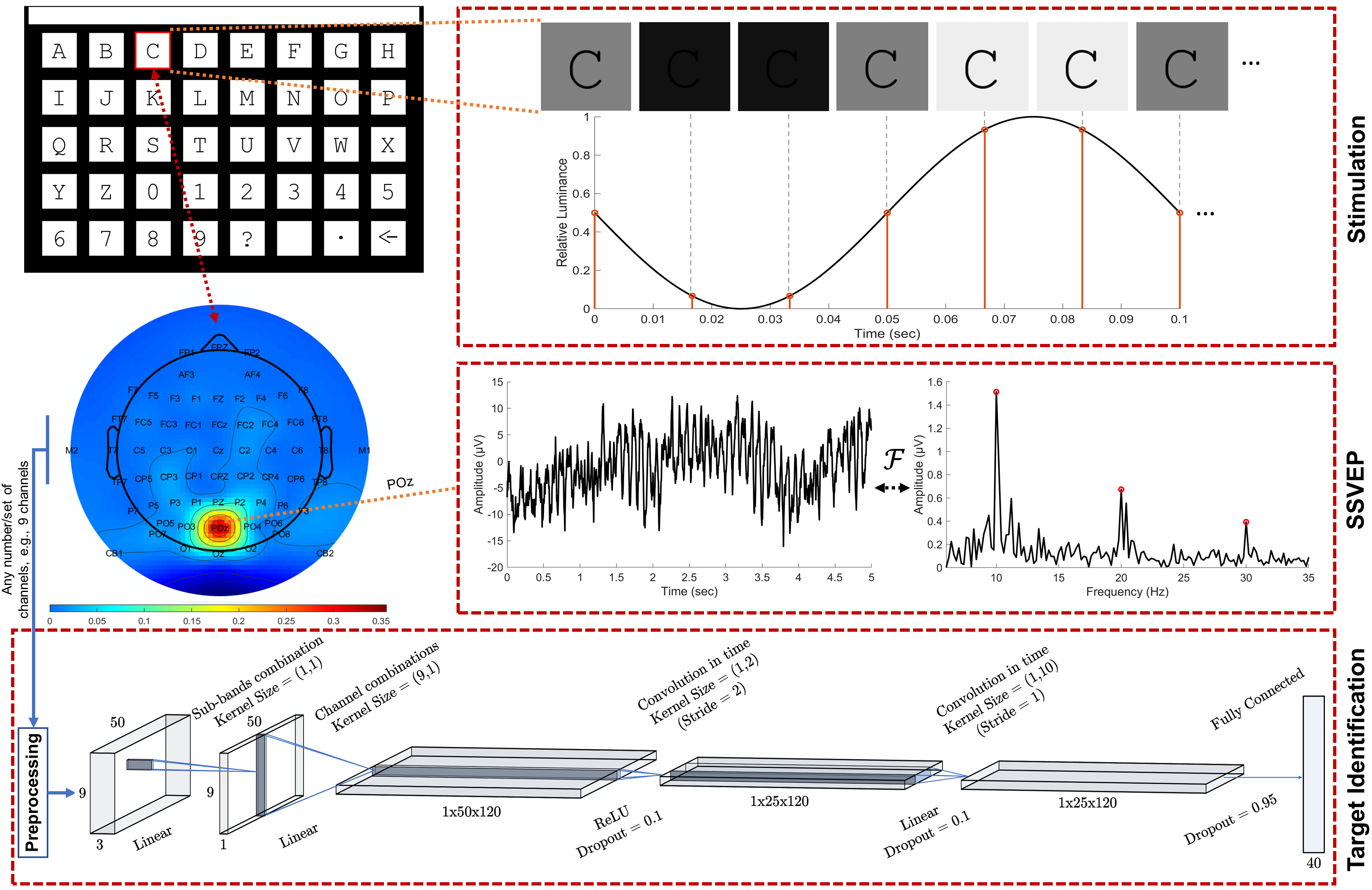}
\centering
\caption{A typical system setup of a BCI SSVEP speller is illustrated. A matrix of thumbnail images of certain alphanumeric characters with IDs $j \in \mathcal{M}=\{1,2, \cdots, M\}$, e.g., $M=40$, is visually presented to the user on the screen. Each character is contrast-modulated in time by a sinusoid of the assigned unique frequency $f_j$, e.g., $f_j\in\{8,8.2,\cdots, 15.8\}$, thereby generating a flickering effect during the $T$ seconds of visual presentation. For example, the character ``C" flickers at $10$ Hz as illustrated above. If the user wishes to spell a character $y \in \mathcal{M}$ and attends to the corresponding thumbnail, then the steady state brain response (when sensed with EEG from particularly the occipital region, cf. the topographic map representing the user) manifests the multi-channel SSVEP signal $x$ that is dominated in its spectrum by the harmonics $\{kf_y\}$ of the input frequency $f_y$, as also illustrated above in the case of $y\equiv$``C". The goal is to predict the target character $y$ based on the received multi-channel SSVEP signal $x$ with $C$ channels. We propose a DNN architecture (with $4$ convolutional and $1$ fully connected layers) for this multi-class classification problem. The proposed DNN strongly outperforms the state-of-the-art and recently proposed techniques uniformly across all signal durations $T\in \{0.2,0.3,\cdots,1.0\}$, but in particular delivers impressive information transfer rate (ITR) results that are $265.23$ bits/min and $196.59$ bits/min in as short as $T=0.4$ seconds of stimulation with $C=64$ channels on the two publicly available benchmark \cite{wang2016benchmark} and BETA \cite{BETA-data} datasets. \textit{To the best of our knowledge, our ITRs are the highest ever reported performance results on these datasets. }}
\label{fig:bcisystem}
\vspace{-3mm}
\end{figure*}

The steady-state brain response to a visual stimulus flickering at a certain frequency induces the SSVEP signal that is characteristically dominated in its spectrum by the harmonics of the applied input flickering frequency. This enables the use of SSVEP in BCI speller designs \cite{bci_speller}. In the experimental paradigm of BCI SSVEP spellers, a matrix of certain alphanumeric characters, each of which flickers at a unique frequency, is presented on the computer screen (Fig. \ref{fig:bcisystem}) and the subject attends to the character that she/he intends to spell. The goal is to predict (i.e. identify) the intended (i.e. targeted) character based on the received SSVEP signal while managing the trade-off between the prediction accuracy and the signal duration such that the maximum ITR is achieved. Since the frequency spectrum is typically exploited up to almost $100$ Hz with the largest harmonic, a high temporal precision and at least $200$ Hz sampling rate are necessary. Hence, EEG is a popular and appropriate choice for its high speed acquisition with a non-invasive and low cost implementation \cite{eegadv}.

We address the target identification in BCI SSVEP speller systems as a multi-class classification problem, and propose a deep neural network (DNN) architecture to the goal of ITR maximization. The proposed DNN processes SSVEP signals in time domain as an end-to-end system from the EEG to the prediction of the targeted character. Our DNN architecture significantly outperforms the state-of-the-art as well as the most recently proposed techniques in the literature. We achieve \textit{impressive ITRs with only $0.4$ seconds of stimulation, $265.23$ bits/min and $196.59$ bits/min,} on the two publicly available large scale benchmark \cite{wang2016benchmark} and BETA \cite{BETA-data} datasets that consist of the EEG data of $105$ subjects with $40$ target characters. \textit{To our best knowledge, these are the highest ever ITRs reported on these datasets.} Moreover, the proposed DNN can be  straightforwardly extended (as it is not specific to spellers) to general BCI systems for the broader purpose of translating brain signals to external commands. Therefore, we believe that our technique will produce a great impact and an immensely valuable use in plethora of real-life applications of SSVEP-based BCIs such as rehabilitation, control and gaming.

\vspace{-2mm}
\subsection{Related Work} \label{sec:rel}
One of the conventional target character identification methods in the literature is based on the power spectrum density analysis (PSDA) of the received SSVEP signal \cite{PSDAmethod}, in which the SNRs of the components of the stimulus frequencies are calculated and then the frequency of the highest SNR is selected as the final prediction. The minimum energy combination (MEC) method \cite{mec_first} linearly combines the SSVEP signals from multiple EEG channels to enhance the identification performance by minimizing the energy of the undesired SSVEP component. Another method is the canonical correlation analysis (CCA) (we call this method as ``Standard-CCA" throughout this paper) in \cite{lin2006frequency}, which
measures the maximal correlation between the SSVEP signal (of the optimal channel combination yielding that maximum) and the reference of a flickering frequency of interest (of the optimal harmonics combination yielding that maximum). Then, the frequency of the largest maximal correlation is selected as the final prediction. Standard-CCA generally demonstrates better ITR performance than PSDA and MEC methods \cite{lin2006frequency,mec-cca-comp}. Aforementioned methods do not incorporate individual data, but incorporating provides a significant ITR performance improvement as noted in \cite{extended-cca2}. 
Therefore, many extensions of Standard-CCA are developed to that end such as ITCCA \cite{it-cca} and the combination of Standard-CCA with ITCCA yielding the method Extended-CCA \cite{extended-cca,extended-cca2}. Among those extensions, Extended-CCA and its improved version m-Extended-CCA are reported to outperform the others \cite{all_cca_comp, cca-jfpm}.


In \cite{filterbank-cca}, running the prediction algorithm in parallel on the multiple SSVEP signals obtained by applying a filter-bank (multiple band-pass filters) first and combining the results afterwards is shown to significantly increase the identification performance of the Standard-CCA method. The reason for this improvement is that the filter-bank approach evaluates the contribution (to the identification) of each harmonic degree separately by using various sub-bands in the spectrum. This is supported in \cite{ssvepsnr} by that, as the degree increases, the harmonic magnitude drops but the SNR does not necessarily decrease since the noise reduces faster. We refer to the seventh figure of \cite{ssvepsnr} for a demonstration, where it is shown that the harmonics up to $50$ Hz maintains a relatively high SNR.  The filter-bank approach has become a standard procedure thereafter and many researchers have followed by utilizing it to increase the performance \cite{cca-jfpm,TS-CORRCA,trca,msTRCA}.

The correlated component analysis (CORRCA) \cite{TS-CORRCA} maximizes the correlation between the multi-channel template signals (which are calculated by averaging the SSVEP signals across multiple trials in the training set for each frequency) and the multi-channel test signal, and then the frequency of the highest correlation yields the final prediction\footnote{\label{myfoot1}Since each character corresponds by design to a unique frequency, we use the phrases ``target character" or ``frequency" and ``identification" or ``prediction" or ``classification" interchangeably depending on the context.} \cite{TS-CORRCA}. The maximization in CORRCA \cite{TS-CORRCA} is with respect to a single projection across channels, whereas the maximization in Standard-CCA \cite{lin2006frequency} is with respect to two projections one of which is across channels and the other is across harmonics in the references. As for the several extensions of CORRCA, the filter bank approach is used in FBCORRCA \cite{TS-CORRCA}, the information from other correlation coefficients is exploited via carefully fusing them  with exponentially decaying weights in HFCORRCA \cite{Hierarchical-CORRCA}, and spatial filters of all stimulus frequencies are utilized in TSCORRCA \cite{TS-CORRCA} yielding the best performing extension.

A method called task related component analysis (TRCA) is used for BCI SSVEP spellers in \cite{trca}. The formulation models the SSVEP signal as a task-related information signal that is linearly contaminated with noise. It is shown in \cite{trca} that TRCA, when used for suppressing the noise in SSVEP by maximizing the inter-trial covariance, delivers higher ITR performance than the Extended-CCA method. TRCA can be enhanced by the filter bank approach along with spatial filters yielding the Ensemble-TRCA (eTRCA) technique \cite{trca}. A multi stimulus scheme (ms-eTRCA) is further incorporated in \cite{msTRCA} which is an advancement over the methods Extended-CCA and eTRCA. 

There exist a few deep learning studies that are related to SSVEP signal classification and BCI spellers \cite{first-deep, bressan2021deep, deep_conference, aznan_dnn, compact_cnn,cnn_ud_ui,convca}. These studies aim to improve the current state with the joint learning of temporal and spatial EEG features via deep neural networks. The joint feature learning not only generates high level representations through cascaded layers but also helps to alleviate the need for a separate preprocessing step. In addition, DNNs allow the inference of nonlinear interactions between such features and the stimulus decoding, which is typically not explored in the conventional techniques. Another theme in this line of research highlights the improvements through subject specific adaptation in the phase of training.

A convolutional neural network (CNN) is designed in \cite{first-deep} to suppress the non-task-related signals in an ambulatory context of SSVEP signal classification. Their network of three layers of multiple feature maps processes the data in the frequency domain and yields a better identification performance than the CCA based methods. The CNN of \cite{bressan2021deep} is composed of temporal and spatial processing layers that are followed by pooling and fully connected layers (FC). It performs favorably compared to the baselines of linear discriminant analysis and random forest in the case of hand movement classification with low frequency EEG (non-SSVEP). A recurrent neural network and a CNN are compared in \cite{deep_conference} against various traditional approaches such as k-nearest neighbor classification, adaboost, decision trees and SVM (together with feature selection), where the CNN (a single convolutional layer, pooling and FC) has been concluded to outperform. The networks of \cite{deep_conference} learns higher level representations starting from power spectral density based EEG features. In contrast, the proposed CNN (a single convolutional layer followed by pooling, batch normalization and FC) in \cite{aznan_dnn} is an end-to-end system (input is raw signal) without preprocessing, and shown to perform better than the approach of \cite{deep_conference} for particularly the dry EEG. All the networks in \cite{first-deep,aznan_dnn,deep_conference,bressan2021deep} are only trained and tested with at most $5$ target classes and $15$ subjects despite that an effective speller requires as many targets as the size of the alphabet. They employ typical activations such as exponential \cite{bressan2021deep} and rectified \cite{deep_conference,aznan_dnn} linear unit (ELU and ReLu), with the exception that the network of \cite{first-deep} is linear. In a different training scheme \cite{compact_cnn}, a compact CNN is trained with the data from a specific training set of subjects and then tested on another that is excluded in the training. This removes the burden of adaptation to a new subject in the phase of deploying a speller system but it is also sub-optimal as the statistics change across individuals. A better strategy is to transfer a pre-trained model and continuously improve it as new personal data becomes available (preferably in an online manner). This idea of fine tuning with transfer learning for subject specific adaptation has been observed in \cite{cnn_ud_ui} to largely improve the identification performance. One further conclusion in \cite{cnn_ud_ui} is that their CNN outperforms the conventional approaches such as CCA, FBCCA and TRCA in both user-dependent and independent settings. Another DNN (Conv-CA) is designed in \cite{convca} for the speller application and reported to deliver a better target identification performance than the method eTRCA \cite{trca}.

Currently, \cite{cnn_ud_ui} and Conv-CA \cite{convca} are the two prominent deep learning studies for SSVEP signal classification. Both of their networks are convolutional. The former \cite{cnn_ud_ui} (spatial and spectral processing layers, batch normalization, ReLu, dropout and FC) receives the magnitude or complex frequency spectrum as the input and trains based on two separate datasets of $21$ subjects with $7$ target classes and $10$ subjects with $12$ target classes. The latter \cite{convca} (three temporal convolutional linear layers, dropout and FC) is deeper with two parallel branches (the signal and reference) outputs of which are correlated in the end for decoding. Also, \cite{convca} considers the speller application with $40$ target classes and $35$ subjects, processes the data in time, and achieves higher ITR. The DNN of \cite{convca} and ours are similar in depth, however, the layers of our architecture are more self explanatory by serving for specific cascaded functionalities such as sub-band combination, channel combination and filtering. In summary, recent reports indicate the efficacy of deep learning for SSVEP signal classification but it is yet to be explored for its full potential.

\vspace{-2mm}
\subsection{Novel Contributions and Highlights}\label{sec:novelcon}
We propose a novel DNN architecture for a BCI SSVEP speller, which processes  SSVEP  signals in  time  domain  as  an  end-to-end  system  from  the  EEG  to the prediction of the target character. The proposed DNN strongly outperforms (with uniformly the highest ITR results for all signal lengths) the state-of-the-art as well as recently proposed deep learning techniques. Our performance evaluations are based on two publicly available large scale datasets that consist of $105$ subjects in total with $40$ target characters. \textit{The code and link to the datasets are available at \url{https://github.com/osmanberke/Deep-SSVEP-BCI}.}

\setlist[itemize]{leftmargin=3mm}
\begin{itemize}
  \item The proposed DNN architecture achieves, with only $0.4$ seconds of stimulation, $265.23$ bits/min and $196.59$ bits/min ITRs on the benchmark \cite{wang2016benchmark} and BETA \cite{BETA-data} datasets. Our ITRs are the highest ever reported performance results on these datasets to date.
  \item EEG signals are well-known to exhibit data statistics that can drastically change from one subject to another in various aspects (e.g. weights of the SSVEP harmonics) but also share similarities in certain other aspects (e.g. SSVEP being tuned to the frequency of stimulation) \cite{train-or-not,towards_zero_training}. To exploit the commonalities while tackling variations, the proposed DNN is trained with two-stages. The first stage trains globally with all the available SSVEP data from all the subjects, and then the resulting DNN model of the first stage is transferred to the second stage that fine-tunes individually to each subject separately.
  \item Most of the techniques in the literature use the conventional $9$ EEG channels (Pz, PO3, PO5, PO4, PO6, POz, O1, Oz, and O2) that are placed over the occipital and parietal regions. Although these channels provide perhaps the most informative SSVEP signals \cite{electrode_position}, other channels might be informative as well. To this end, our proposed DNN is fed with all the available channels and learns an optimal combination of the channels; and this is even for each subject separately taking into account that the optimal combination can change across subjects. In fact, we observe a significant improvement (by about $20$ bits/min on the benchmark dataset \cite{wang2016benchmark} and $10$ bits/min on the BETA dataset \cite{BETA-data}) in ITR compared to the conventional pre-determined set of $9$ channels. Therefore, a manual channel selection or running an independent algorithm for that is not required in our method.
  \item Unlike the other techniques (e.g., \cite{filterbank-cca,Hierarchical-CORRCA}), neither the EEG channel selection nor the harmonics combination in our method is manual or rule based. Rather, we jointly learn the both with the proposed DNN in a completely data driven manner. Optimizing the sub-band combination is important since the harmonics can have significantly different magnitudes and contributions to the target identification. For instance, higher order harmonics having low magnitude does not necessarily mean that they have low contribution, since the SNR might be high despite the low magnitude (cf. \cite{ssvepsnr} and also our Fig. \ref{fig:bcisystem}). Hence, a normalization across frequency bands via a weighted sub-band combination is implemented in the first layer of our DNN. This improves the target identification accuracy by about $2\%$ in the benchmark dataset \cite{wang2016benchmark} and $2.5\%$ in the BETA dataset \cite{BETA-data}, at  $0.4$ seconds of signal acquisition, where the chance level is $1/40=2.5\%$. 
 
\end{itemize}

\noindent
In the following Section \ref{sec:PR} and Section \ref{sec:PS}, we provide the problem description and the proposed DNN as our solution. After we present the performance evaluations in Section \ref{sec:EX}, the paper concludes in Section \ref{sec:CO} with final remarks. 
\vspace{-1mm}
\section{Problem Description} \label{sec:PR}
\normalsize {
During a trial in a BCI SSVEP speller session (illustrated in \autoref{fig:bcisystem}), the subject is visually presented a matrix of $M$ alphanumeric characters each of which flickers at unique frequency $f_j:j\in\mathcal{M}=\{1,2,\cdots,M\}$ (in Hz), e.g., $f_j\in\{8,8.2,\cdots, 15.8\}$ with $M=40$. Then, she/he is asked to concentrate on the target character with the identification number $y\in \mathcal{M}$ that is to be spelled. The brain response, as a result of the stimulation by the intended target character $y$ flickering at the frequency $f_y$, is measured with EEG as the multi-channel SSVEP signal $x\in \mathbb{R}^{C \times N}$. Here, $C$ is the number of channels and  $N=T\times f_s$ is the number of samples in each channel (with $T$ and $f_s$ being the signal or stimulation duration in seconds and the sampling frequency in Hz, respectively). We emphasize that a BCI SSVEP speller system is typically designed for enabling a severely motor disabled individual to \textit{communicate flawlessly at a fast rate which requires a high speed accurate speller}. Therefore, the main design goal is to maximize the ITR \cite{itr_cite} that is a function of the target identification accuracy and the stimulation duration.  If  the prediction is perfectly accurate, then the trial-by-trial spelling of a length-$l$ word requires $T\times l$ seconds which is equivalent to the ITR $\log_2 M \frac{60}{T}$ bits/min, i.e., }

{\small {
\begin{align}
    \text{ITR}(P,T) &= (\log_2M + P\log_2P + (1-P)\log_2 \left[ \frac{1-P}{M-1}\right])\frac{60}{T} \nonumber \\
    &=(\log_2 M) \frac{60}{T} \text { (when $P=1$) }.
\end{align}}}
\normalsize
Note that the prediction accuracy $0\leq P \leq 1$ is almost never perfect; nevertheless, if the identification method is optimal (with the minimum possible error rate, i.e., $1-P$), then the $P$ can be improved only by requesting a longer stimulation via increasing $T$ (resulting in a larger amount of data). However, in this case of lengthening the stimulation duration, the trials of the spelling slow down and consequently the ITR does not necessarily improve. For example, the long stimulation $T=\infty$, results in the perfect accuracy $P=1$ that is, though, $0$ ITR.  Hence, when the identification method is optimal, it is not possible to expedite the spelling while also improving the $P$ since the two are incompatible. This requires to manage a trade-off between $P$ and $T$ for the ITR maximization. On the other hand, when the target identification is itself not optimal, improving the $P$ is possible without increasing the $T$ up to the point where the trade-off starts dominating.

In this paper, we not only manage the trade-off between the $P$ and the $T$ but also obtain the optimal target identification for the ITR maximization. Accordingly, we formulate the character identification as a multi-class classification problem based on the data $\{(x_i,y_i)\}_{i=1}^D$, where $D$ is the number of trials, to the goal of designing a classifier $g(x)=\hat{y}$ such that the ITR is maximized. Since the ITR maximization for a fixed $T$ is equivalent to accuracy maximization, our strategy is to minimize the $1-P$ for each $T$, and observe the pair $(P^*, T^*)$ that yields the maximum ITR.
\vspace{-1mm}
\section{Proposed Solution: A DNN Architecture} \label{sec:PS}
In the following, we first briefly explain the SSVEP signal characteristics, and then describe the proposed DNN architecture while also motivating its main design components.
\vspace{-2mm}
\subsection{The SSVEP Signal}
The stimulus in BCI SSVEP spellers characteristically leads to the SSVEP signal $x$ that mostly comprises of the frequency components $A_f\cos(2\pi ft + \phi_f)$ (where $t=n/f_s$ due to sampling) at the harmonics $f=kf_y$ (integer $k$) of the stimulation frequency $f_y$. The entire spectrum (up to typically $100$ Hz as far as the information content, which is corrupted by the noise and interfered with other ongoing processes in the brain, is concerned) is spanned, but the components of the harmonics are larger, i.e., $A_{kf_y}>>A_f>0$ for $f\neq kf_y$ \cite{bci_speller,cca-jfpm} (cf. also the spectrum example in Fig. \ref{fig:bcisystem}). Then the target identification problem in this setting can be perhaps solved by the detection of the peaks across harmonics up to a certain degree in the Fourier spectrum of the SSVEP signal. Namely, one can decide for the character whose harmonics are most covered by the spectrum. However, the harmonics are generally not observable in the spectrum as orthogonal components since the signal duration $T$ yields only a low frequency resolution $\delta \hat{w}=\frac{2\pi}{Tf_s}$ rad in normalized radian frequency and $\delta f = \frac{\delta \hat{w}}{2\pi}f_s=\frac{1}{T}$ Hz in cyclic frequency (where $T$ is short). Therefore, the information in the harmonics of $A_{k\delta f} \cos(2\pi k\delta ft + \phi_{k\delta f})$ (where $t=n/f_s$ due to sampling) do leak onto the entire Fourier spectrum of the SSVEP signal due to the correlation between the harmonics and the spectrum components. Thus, several variants of the CCA can be used to find the flickering frequency $f_{\hat{y}}:\hat{y}\in\mathcal{M}$, which yields the maximum available correlation between the optimal linear combination across channels in the received SSVEP signal and, for instance, A) the optimal linear combination of the corresponding reference harmonics, e.g., \cite{lin2006frequency,cca-jfpm}, or B) the optimal linear combination across channels in the template SSVEP signals in the training set, e.g., \cite{Hierarchical-CORRCA,TS-CORRCA}. This is currently the essential principle in the state-of-the-art techniques. We refer to Section \ref{sec:rel} for a detailed discussion.

Despite being certainly impressive, such state-of-the-art techniques have issues in various aspects. This presents an opportunity for a performance improvement (once those issues are addressed). For example, the application of the CCA in its current form of those techniques essentially measure the similarity between a test signal and each class mean of the respective training signals for obtaining a decision. This is sub-optimal as measuring the similarities to each training signal (rather than only mean) and then basing the decision on all such measured similarities would be a better approach (this resembles the difference between the nearest neighbor classifier and the nearest mean classifier). Also, CCA returns only a linear model whereas we observe in our performance evaluations that a nonlinear model is in fact needed. To this end, in the following, we describe the proposed DNN while explaining how especially certain issues (the one mentioned above is only an example among many others) we have identified in the literature are resolved.
\vspace{-2mm}
\subsection{The Proposed DNN Architecture}
Our DNN architecture operates as an end-to-end system which receives the multi-channel SSVEP signal $x$ and processes it in a feed-forward manner to the final prediction $\hat{y}$. The proposed DNN (\autoref{fig:bcisystem}) consists of $4$ convolutional layers and $1$ fully connected layer. Hence, we have the processing $x \rightarrow \text{preprocessing: } [x^{(1)},\cdots,x^{(r)}, \dots, x^{(N_s)}] \rightarrow z_1 \rightarrow z_2 
    \rightarrow z_3 \rightarrow z_4 \rightarrow s \rightarrow \hat{y}=\arg\max{s(j)},$
where the preprocessing is for generating the sub-bands of harmonics $[x^{(1)},\dots, x^{(N_s)}]$ that are combined in the first layer to produce the $z_1$ which is processed in the second layer for spatial filtering  to produce the $z_2$. Here, $N_s$ is the number of sub-bands and $r$ is the corresponding index. Downsampling follows in the third layer, yielding $z_3$, then features are extracted in the fourth layer as $z_4$ passing to the classification in the fully connected layer to produce the prediction $\hat{y}=\arg\max{s(j)}$ ($s\in [0,1]^{M\times 1}$ is the softmax output). 

{\bf Remark: }We point out that since there is only one nonlinear activation (ReLu) in the proposed DNN, it can be reduced to a single hidden layer network with the ReLU activation at the hidden layer. However, that would only lead to a non-intuitive complicated network. In our DNN, the information flow is natural through an intuitive and conceptually simple design. Therefore, we present the  DNN as the composition of $5$ functionalities, i.e., layers.

In the following, we describe our proposed DNN. For each layer, we first motivate its use and then provide its definition. Next, the training scheme and the further details are explained.     

\subsubsection{First layer for harmonics (sub-bands) combination}
Under the stimulation by the target character flickering at the frequency $f_y$, the contributions of the harmonics to the generation of the SSVEP signal might vary from one harmonic to another. For example, a lower order harmonic generally has a larger magnitude compared to a higher degree one \cite{cca-jfpm}. Nevertheless, since the higher order harmonics tend to be less (for example, compared to the alpha band around $10$ Hz) interfered with other ongoing brain activities, they tend to manifest perhaps surprisingly a relatively high SNR \cite{ssvepsnr} (as we also observe by inspection in the spectrum example of Fig. \ref{fig:bcisystem} in the case of the stimulation frequency $10$ Hz up to the $3$rd degree). We also refer to the study \cite{ssvepsnr} for a general SNR investigation of SSVEP harmonics.  However, it is not straightforward to assess which harmonic is more informative in the SSVEP classification, and hence how to normalize in the spectrum across the harmonics could be a fairly difficult task. This issue is handled in the literature by processing several sub-bands of the SSVEP spectrum separately, but then the results are fused in a rule based manner or are fused based on a fairly restrictive model, e.g., \cite{filterbank-cca,cca-jfpm}. Therefore, how to choose the weight of a certain harmonic is not sufficiently addressed in the literature due to their manual handling.

In our DNN design, we opt to stay agnostic about this normalization of harmonics and instead let the network decide about the normalization weights by training in a  data driven manner. For this purpose, we band-pass filter (denoted by $\mathcal{G}_r$, with MATLAB filtfilt function) the SSVEP signal $x\in \mathbb{R}^{C\times N}$ in each channel (multiple times $1\leq r\leq N_s$), where the lower cut-off is $r\times\min\{f_j\}-\epsilon$ Hz (e.g., $\sim 8$ Hz for $r=1$ in both of the datasets \cite{wang2016benchmark,BETA-data}) and the upper cut-off is $6 \times \max\{f_j\}+\epsilon$ Hz (e.g., $\sim 90$ Hz in both of the datasets \cite{wang2016benchmark,BETA-data}) with $\epsilon$ being a small margin. The filter is designed as (using MATLAB designfilt function) zero-phase Chebyshev-Type 1 with filter order 2 and 1 dB pass band ripple. Hence, each filter $\mathcal{G}_r$ excludes the harmonics of the degree that is less than $r$ while including the rest up to the $6$'th degree (the maximum degree is set to $6$ since beyond $100$ Hz in the EEG is typically noise in BCIs). This yields the filtered output $x^{(r)}\in \mathbb{R}^{C\times N}$ that includes a specific sub-band of harmonics. 

The first layer of our DNN (with the weights $w_1\in\mathbb{R}^{N_s \times 1}$) linearly combine these sub-bands for a normalization across the harmonics as $z_1=[x^{(1)},\cdots,x^{(r)}, \dots, x^{(N_s)}]w_{1}$, where the input to the layer is $[x^{(1)},\cdots,x^{(r)}, \dots, x^{(N_s)}]\in\mathbb{R}^{C \times N \times N_s}$ (i.e. a volume of $9\times50\times3$ in the case of $C=9$, $N=50=T\times f_s$ with $T=0.2$ seconds, $f_s=250$ Hz, and $N_s=3$) whereas the output is $z_1 \in\mathbb{R}^{C\times N}$ (i.e., a plane of $9\times50\times 1$ when $C=9$ and $N=50$).
Hence, (if desired) our DNN has the capability to amplify the higher order harmonics  by choosing the corresponding weights relatively high. 

\subsubsection{Second layer for channel combination}
The SSVEP signal is a multi-channel signal. The channels, on the one hand, bear valuable distinct information from the brain regions that they sense from but, on the other, produce signals that are also largely correlated. A combination across channels shall be considered to extract and accumulate the whole information while discarding the redundancy or non-informative variations. Also, multiple combinations are probably needed since extracting the information living in one subspace (of the complete channel space) can suppress the one living in another subspace. The required combinations might be even more than the number of channels as those informative subspaces are not necessarily orthogonal, requiring in return a nonlinear processing of combinations. The existing CCA analyses in the literature (such as \cite{cca-jfpm,TS-CORRCA}) allows a separate channel combination for each and every single test instance. Here, we criticise this since it not only A) creates an unnecessarily large degree of freedom and in return a large detrimental effect due to the induced strong proneness to overfitting, but also B) risks suppressing, in each case of the test instances, valuable information that can be extracted by other but not utilized combinations. To alleviate the issue B, those techniques incorporate multiple combinations -for each and every testing again- by fusing the correlation coefficients of the CCA analysis, but this further worsens the issue A. Even then, the number of combinations is limited by the number of channels due to linearity, and the coefficient fusing is typically rule-based without a data-driven learning or is based on a simple fitting to a rather restrictive model. Further, a regularization step is generally not incorporated though it is certainly needed.

In the following, we explain from another perspective to motivate our approach. Since the neural circuitry and nonlinear processes that are involved in brain to generate the SSVEP responses vary from lower degree harmonics (as well as intermodulations) to upper degree ones \cite{neuralcircuit}, we certainly expect that different brain regions are more responsive to different harmonic degrees which necessitates the use of multiple channel combinations. Moreover, a different combination might be more appropriate to emphasize a certain stimulation frequency and its harmonics while suppressing the others which further necessitates multiple combinations for each classes.

Unlike the state-of-the-art methods, in our DNN design, we use the same set of multiple channel combinations that is common for all of the instances. This set in our study includes as many combinations as the number $N_s$ of sub-bands for each stimulation frequency $f_j$, yielding in total, for instance, $N_{ch}=120=N_s\times M$ combinations when $N_s=3$ and the number of characters is $M=40$. We emphasize that if the number $N_{ch}$ of channel combinations is more than the number $C$ of channels, then one needs nonlinear processing (to avoid degeneracy and) to make use of the combinations effectively. Overall, this setting keeps the parameter complexity at a manageable level and mitigate the overfitting when compared to using a separate combination for each and every single test instance as in the existing techniques of literature. At the same time, our setting is also sufficiently powerful since we can use combinations as many as needed. To this end, the second layer (parameterized over the weights $w_2\in\mathbb{R}^{C \times N_{ch}}$) of our DNN combines the channels by receiving the input plane $z_1 \in\mathbb{R}^{C\times N}$ and returning the plane $z_2\in\mathbb{R}^{N\times N_{ch}}$, i.e., $z_2=z_1'w_2$, where $z_1'$ is the transpose of $z_1$. In order to achieve nonlinearity, we also apply the ReLU (rectified linear unit) activation but postpone it until the end of the third layer.

\subsubsection{Third layer for filtering in time, downsampling and nonlinearity}
This layer has two functions. First, it applies a filter of size $2$ in time (with also a full third dimension along the depth) with stride $2$, thereby halving the dimension (downsampling by $2$) and reducing the parameter complexity in the network. This operation can be considered to represent the anti-aliasing filtering that is commonly used with downsampling. The filtering in this layer additionally serves for roughly adjusting the spectral bandwidth for each information flow over the channel combinations in the network. Hence, multiple such filters (as many as the number of channel combinations, i.e. $N_{ch}=120$) are used. Second function of this layer is applying the nonlinearity. Note that when we have $N_{ch}>C$, the input plane of this third layer $z_2\in\mathbb{R}^{N\times N_{ch}}$ is rank-deficient with a rank at most $C$ even if $z_1 \in\mathbb{R}^{C\times N}$ (producing $z_2=z_1'w_2$) is full rank. This defeats the purpose of producing multiple channels combinations in the previous layer. Hence, to tackle the rank-deficiency and enable the effective use of the channel combinations, the nonlinear ReLU activation is applied after downsampling to produce the output plane $z_3\in\mathbb{R}^{(N/2)\times N_{ch}}$. 

\subsubsection{Convolutional fourth and fully connected fifth layers}
The fourth convolutional layer filters the input $z_3$ with multiple finite impulse response filters (FIRs, each being of length $10$ with also a full third dimension along the depth) to produce the features in $z_4$ that is finally classified by the following fully connected (FC) layer. Hence, the very first input $x$ is predicted as $\hat{y}=\arg\max{s(j)}$ ($s\in [0,1]^{M\times 1}$ is the softmax output of the FC layer). The FIR filters in the fourth layer are expected to achieve frequency responses that are tuned to the spectral patterns of each stimulation class ($1$ FIR for each sub-band per each $M=40$ classes, yielding in total $120$ filters when we have $N_s=3$ sub-bands) for extracting powerful features.  
Hence, in these two layers, all of the FIR filters as well as all of the FC weights are optimized.

\subsubsection{Two-staged training and further details}
The proposed DNN is initialized by sampling the network weights from the Gaussian distribution with $0$ mean and $0.01$ variance, except that all of the weights in the first layer are initialized with $1$'s. The exception of the first layer is due to an intuitive choice for assigning equal weights to the sub-bands initially without affecting the order of magnitudes of the input filtered signals. We train the network in each iteration based on the training batch data $\{(x_i,y_i)\}_{i=1}^{D_b}$, where $D_b$ is the number of trials in the batch, by minimizing the categorical cross entropy loss
\begin{align}
    \frac{1}{D_b}\sum_{i=1}^{D_b} -\log(s_i(y_i)) + \lambda |\mathbf w|^{2} 
\end{align}%
via the Adam optimizer \cite{kingma2014adam} with the learning rate $\nu=0.0001$ (without decaying), where $\lambda$ is the constant of the L2 regularization which we set as $\lambda=0.001$, $s_i\in[0,1]^{M\times1}$ is the softmax output for the instance $x_i$, $s_i(y_i)$ is the $y_i$'th entry of $s_i$ and the final prediction is  $\hat{y}_i=\arg\max s_i(j)$. Here, $\mathbf w$ represents all the DNN weights.
Dropout layers are incorporated between the second and third, third and fourth, and fourth and fifth layers with dropout probabilities $0.1$, $0.1$, and $0.95$, respectively. 

We also point out that the total number of trainable parameters in our proposed DNN can be found by $N_s + C  N_{ch}+ 2  N^2_{ch} + 10  N^2_{ch} + \frac{N}{2}  N_{ch}  M$ (each term is for a layer including the output layer), which yields $413,883$ parameters in the plausible setting of $N_s=3, C=9, N_{ch}=120, T=0.4 \text{ sec}, f_s=250 \text{ Hz}$ with $M=40$. Since there are at most $8400$ training instances in the considered datasets, which is low compared to the parameter complexity, we opt for a relative strong regularization by using a large dropout probability ($0.95$) in the last layer.

We train the network in two stages. The first stage takes a global perspective by training with all of the data (in the training set) whereas the second stage re-initializes the network with the global model and fine-tunes it to each subject separately by training with only the corresponding subject data (of the training set). Hence, each subject has her/his own model. Most of the existing studies do either develop only a local model (e.g., \cite{trca,TS-CORRCA}) or only a global model (e.g., \cite{first-deep}), which indicates that our introduced two-stage training is also a novel contribution to BCI SSVEP spellers (cf. Sections \ref{sec:rel} and \ref{sec:novelcon}). We observe that this idea of transfer learning with two-staged learning, since it takes into account the inter-subject statistical variations, provides significant ITR improvements. In the following section of the performance evaluations, we study with two datasets independently. Namely, the global model of the first stage training is obtained for each dataset separately rather than training a single global model based on the union of the two datasets.

\vspace{-2mm}
\section{Performance Evaluations} \label{sec:EX}
We test our DNN on publicly available two datasets which are the benchmark \cite{wang2016benchmark} and the BETA \cite{BETA-data} datasets. The state-of-the-art techniques have been previously tested on these datasets; and in our evaluations, we compare against specifically those that have been reported to perform well. In particular, we compare against $7$ methods: Conv-CA, ms-eTRCA, eTRCA, TSCORRCA, m-Extended-CCA, Extended-CCA and CORRCA. In our comparisons, we follow the same test procedures for all these methods.      

A BCI SSVEP speller experiment consists of several blocks, so that the subject can have a break between two blocks. For example, there are $6$ and $4$ blocks in the benchmark and BETA datasets, respectively. In our performance evaluations, we conduct the comparisons (following the same procedure in the literature) in a leave-one-block-out fashion. We train on $5$ (or $3$) blocks and test on the remaining one and repeat this process $6$ (or $4$) times  to test on each block in the benchmark (or the BETA) dataset. For each signal duration $T$ in the range $T\in \{0.2,0.3,\cdots,1.0\}$, we report the mean classification accuracy and ITR along with the standard errors. We take into account a $0.5$ seconds gaze shift time while computing the ITR.  We test with the pre-determined set of $9$ channels (Pz, PO3, PO5, PO4, PO6, POz, O1, Oz, and O2) again for fair comparisons since these channels have been used in the compared methods, but we also test with all of the available $64$ channels to fully demonstrate the efficacy of our DNN. In fact, we observe improvements with $64$ channels over the pre-determined set. Confusion matrices are also presented for further insights into our classification results. Additionally, we analyze the effect of the number of sub-bands and channels on the identification performance. We also report the topographic channel distributions to demonstrate the weight of each channel's contribution to the our DNN performance. 

\begin{figure}[t]
\includegraphics[width=0.28\textwidth]{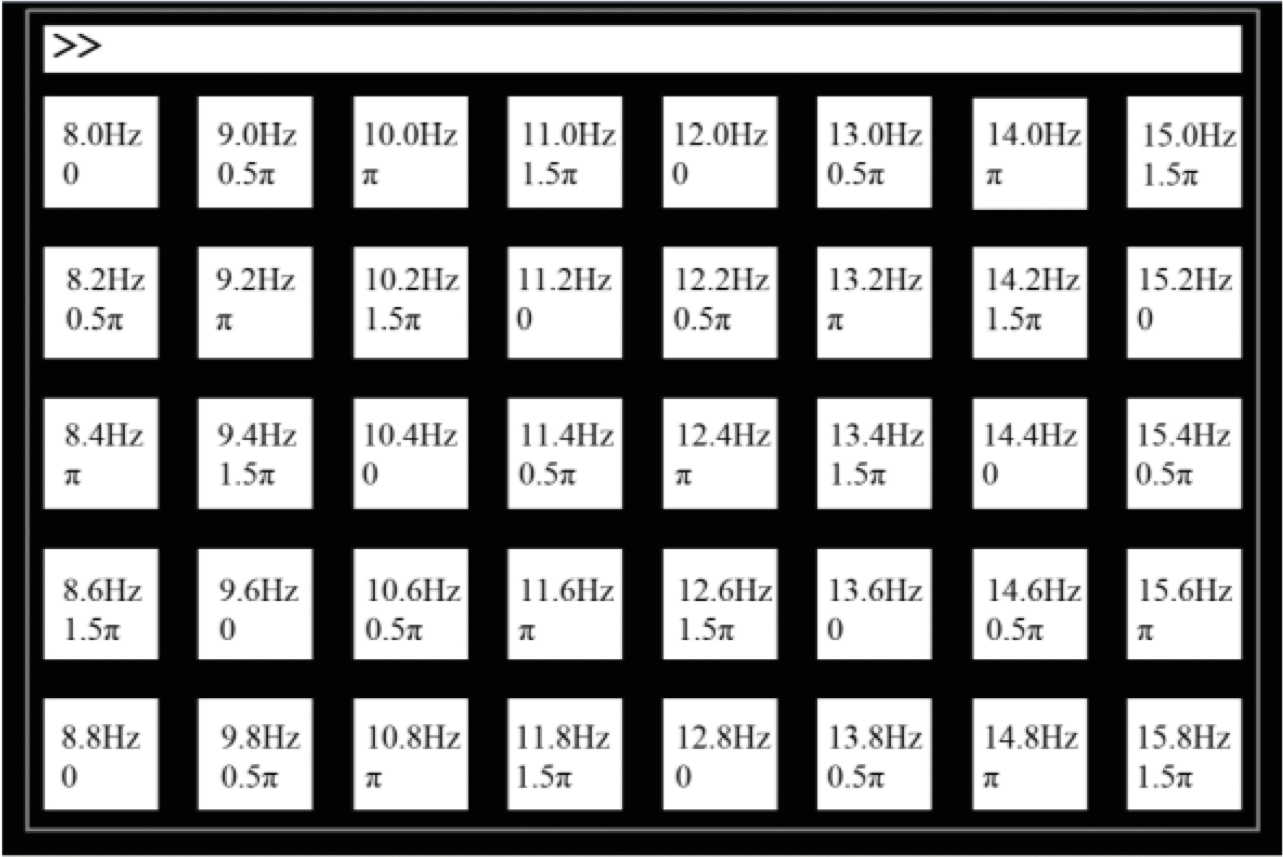}
\centering
\caption{The character matrix layout for the stimulus presentation in the experiments of the benchmark dataset is shown (image is taken from \cite{wang2016benchmark}).}
\label{fig:bench_screen}
\end{figure}
\begin{figure}[t]
\includegraphics[width=0.28\textwidth]{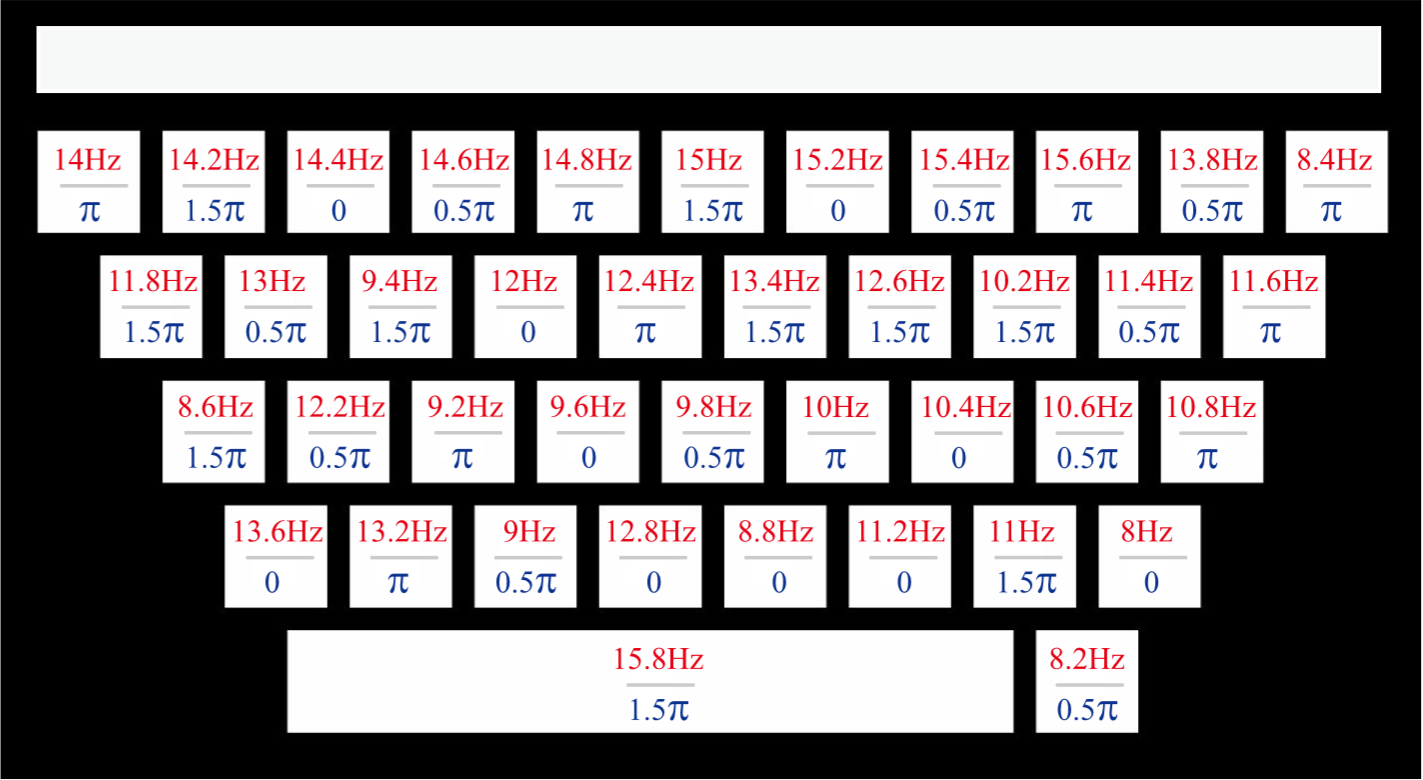}
\centering
\caption{The character matrix layout for the stimulus presentation in the experiments of the BETA dataset is shown (image is taken from \cite{BETA-data}).}
\label{fig:beta_screen}
\vspace{-3mm}
\end{figure}

{\bf The benchmark dataset} has been recorded in BCI SSVEP speller experiments with $35$ healthy subjects. Each experiment consists of $6$ blocks, i.e., sessions. During a block, the subject is shown on the screen (\autoref{fig:bench_screen}) a matrix ($5 \times 8$) of $40$ target characters flickering at various frequencies (in the range $8-15.8$ Hz with $0.2$ Hz increments) with at least $0.5 \pi$ phase difference between adjacent frequencies. The EEG data is recorded through $64$ channels. Each block includes $40$ random-order trials (one trial per each target character). Each trial starts with a visual cue that is displayed for $0.5$ seconds on the screen to guide subject's gaze to the desired target, and then conducts the stimulation for $5$ seconds that is followed by an offset of $0.5$ seconds. The EEG is downsampled to $250$ Hz. Average visual latency of the subjects is approximately estimated as $140$ ms in this dataset. We refer to
\cite{wang2016benchmark} for further details.

\begin{figure}[t!]
    \centering
    \includegraphics[width=0.47\textwidth]{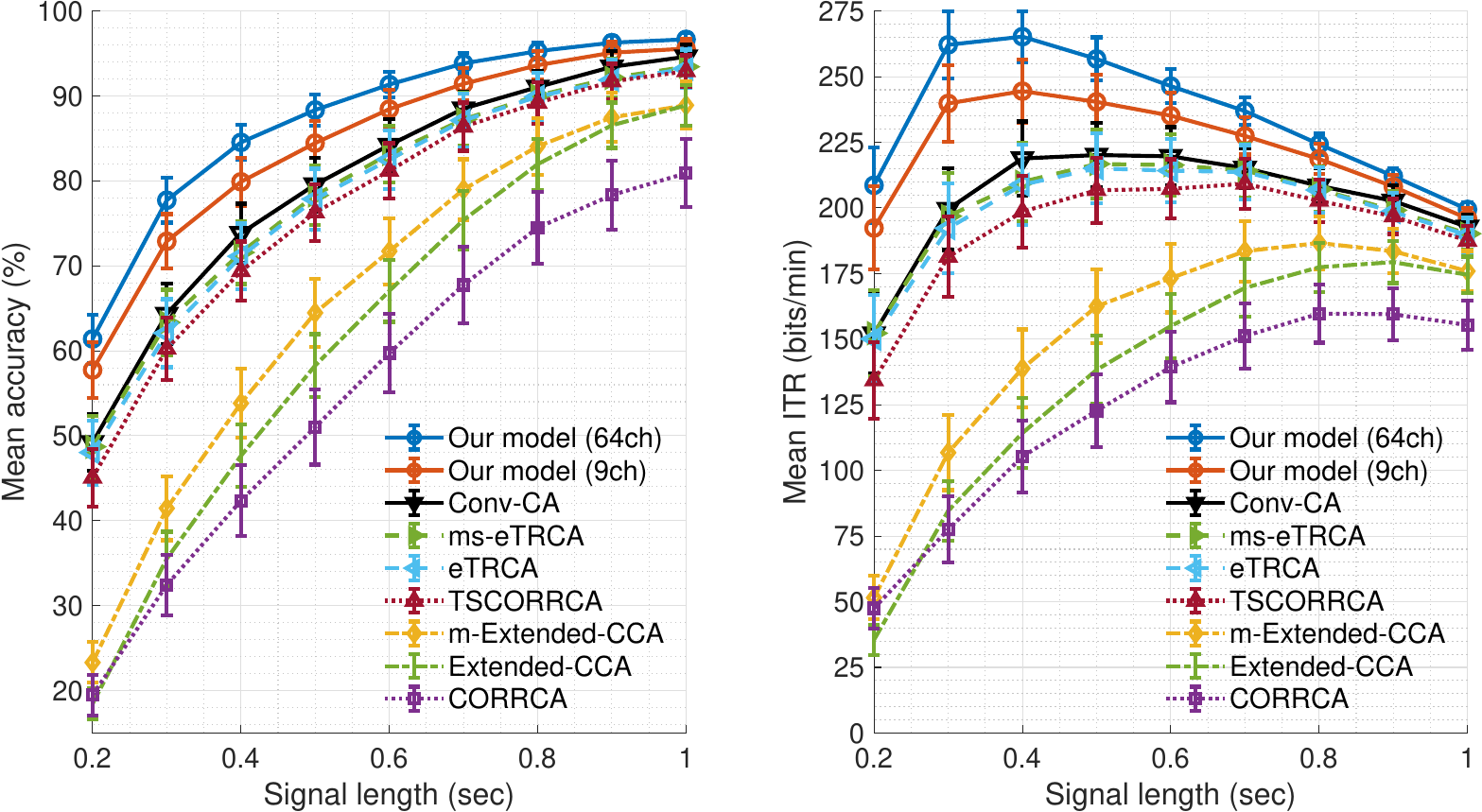}
    \caption{The mean classification accuracy on the left and the mean information transfer rate (ITR) on the right are presented across all $35$ subjects in the benchmark dataset, together with the standard errors indicated by the bars.}
    \label{fig:bench_res}
    \centering
    \vspace{-3mm}
\end{figure}

{\bf The BETA dataset} and the benchmark dataset are similar, but also have certain important differences. We note the differences in the following (the remaining attributes are the same). This BETA dataset has been recorded with $70$ healthy subjects. Each experiment consists of $4$ blocks. The flickering target characters are shown in the form of a keyboard (\autoref{fig:beta_screen}). The experiments are conducted outside of the laboratory environment, resulting in a lower SNR compared to the benchmark dataset. Hence, the target identification is more challenging in this case. The stimulation lasts $2$ seconds for the first $15$ subjects and $3$ seconds for the remaining subjects. Average visual latency of the subjects is approximately estimated as $130$ ms in this dataset. We refer to
\cite{BETA-data} for further details.


Since the available data shrinks in the second stage of our DNN training, to achieve a better regularization, the probabilities of the first two dropout layers are increased to $0.6$ for the benchmark dataset \cite{wang2016benchmark} and to $0.7$ for the BETA dataset \cite{BETA-data}. A larger dropout probability is used for the BETA dataset as it is smaller in size (per subject) and more noisy. The number of epochs (without early stopping) are $1000$ and $800$ in the first stage for the benchmark dataset and the BETA dataset, respectively, where the batch size is $100$ for the both. In the second stage, the number of epochs (without early stopping) are the same and $1000$ for both of the datasets and the batch sizes are $200$ and $120$ for the benchmark dataset and the BETA dataset, respectively. All the other settings of the proposed DNN are exactly the same between the two stages and also between the two datasets.

\vspace{-2mm}
\subsection{Results}
\begin{figure}[t!]
\includegraphics[width=0.47\textwidth]{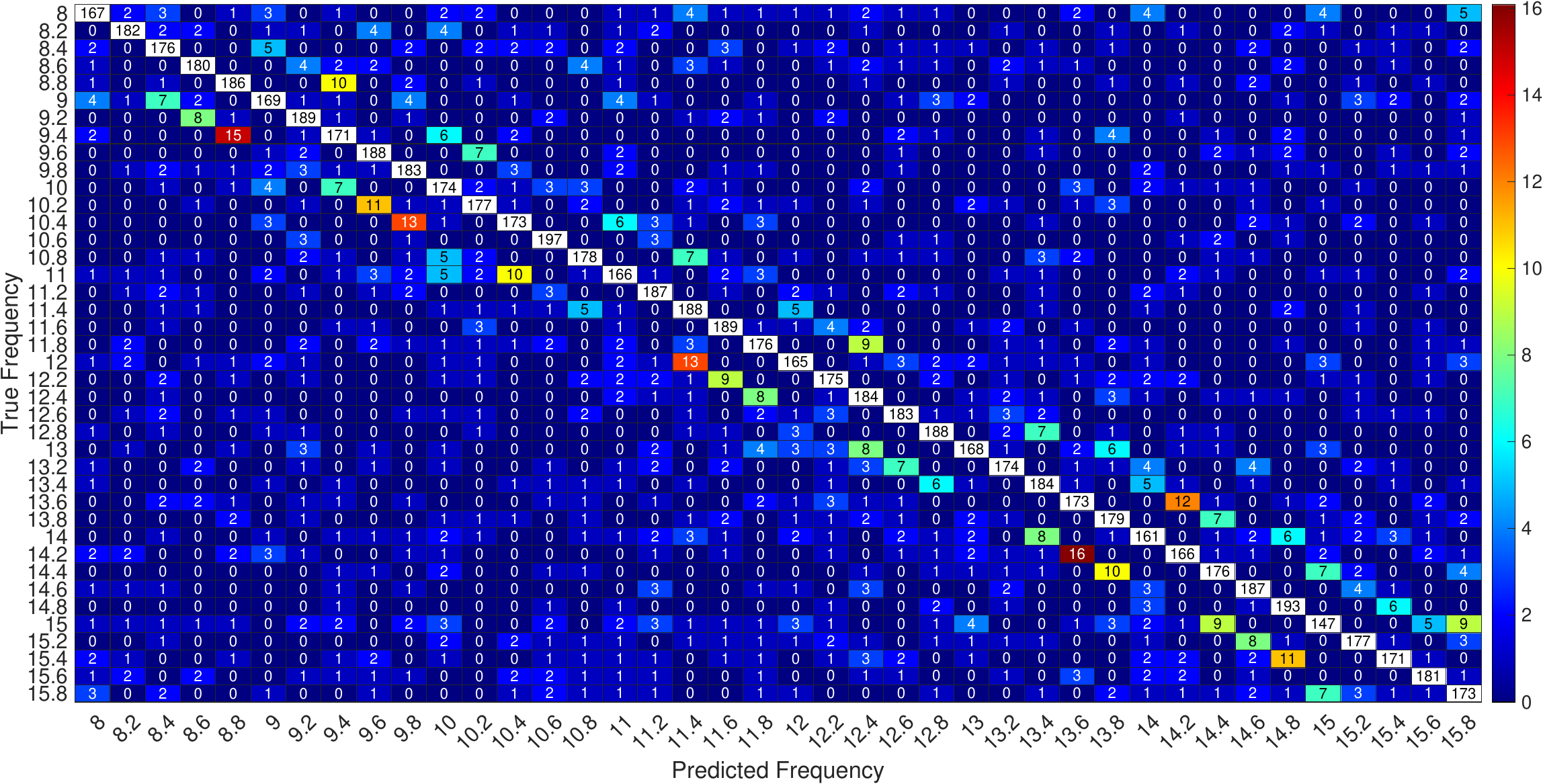}
\centering
\caption{The confusion matrix of the proposed DNN with $64$ channels on the benchmark dataset  at $0.4$ seconds of stimulation.\protect\footnotemark}
\label{fig:bench_conf}
\vspace{-2mm}
\end{figure}

\begin{figure}[t!]
    \centering
    \includegraphics[width=0.47\textwidth]{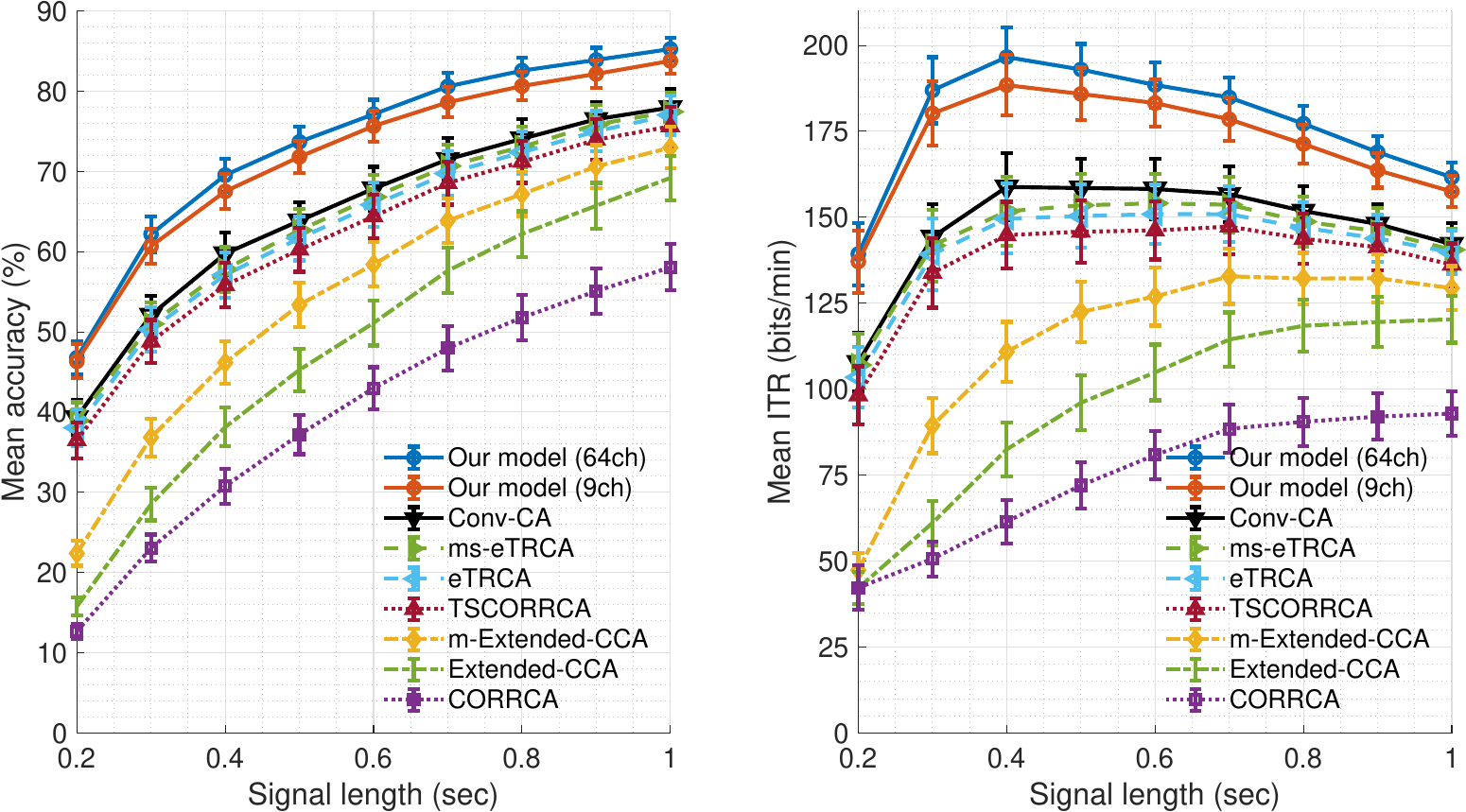}
    \caption{ The mean classification accuracy on the left and the mean information transfer rate (ITR) on the right are presented across all $70$ subjects in the BETA dataset, together with the standard errors indicated by the bars.}
    \label{fig:beta_res}
    \centering
    \vspace{-3mm}
\end{figure}

\begin{figure}[t]
\includegraphics[width=0.47\textwidth]{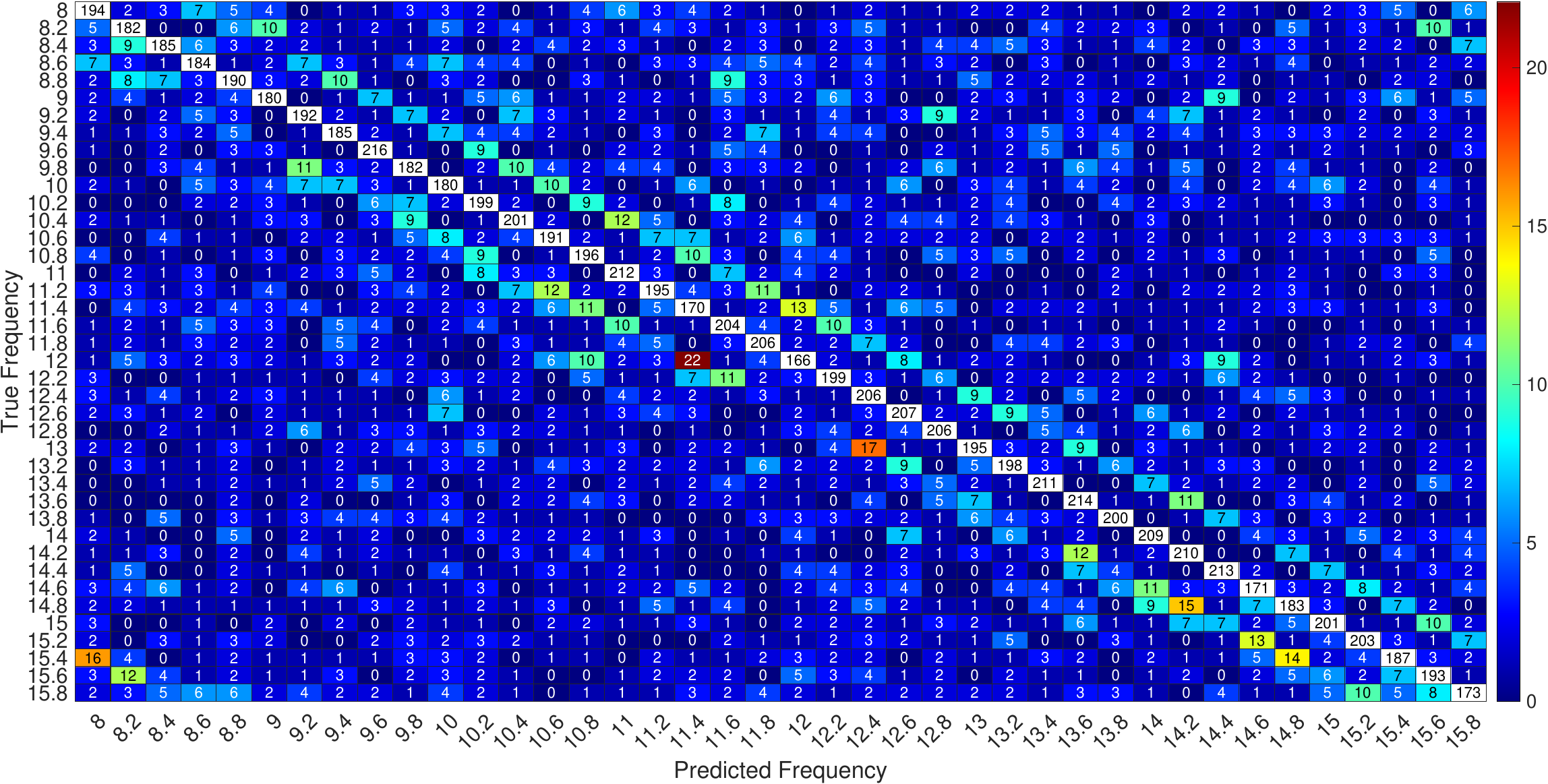}
\centering
\caption{The confusion matrix of the proposed DNN with $64$ channels on the BETA dataset at $0.4$ seconds of stimulation\textsuperscript{2}}
\label{fig:beta_conf}
\vspace{-2mm}
\end{figure}

The proposed DNN is observed to achieve $265.23$ bits/min ($\sim84\%$ accuracy with $64$ channels) and $244.45$ bits/min ($\sim80\%$ accuracy with $9$ channels) maximum ITRs (cf. \autoref{fig:bench_res}, and the corresponding confusion matrix in \autoref{fig:bench_conf} with $64$ channels and $0.4$ seconds of stimulation) on the benchmark dataset, and $196.59$ ($\sim70\%$ accuracy with $64$ channels) bits/min and $188.45$ ($\sim68\%$ accuracy with $9$ channels) bits/min on the BETA dataset (cf. \autoref{fig:beta_res}, and the corresponding confusion matrix in \autoref{fig:beta_conf} with $64$ channels and $0.4$ seconds of stimulation). These results are achieved in only $T=0.4$ seconds of stimulation with using $N_s=3$ sub-bands. \textit{The fact that we observe such impressive results with the same exact setting in both of these independent datasets is particularly important and thereby providing further reassurance about the robustness of our presented results.} In fact, across all stimulation durations, the proposed DNN strongly outperforms all the other techniques in terms of both the accuracy and ITR, in both datasets.

In the rest of our performance evaluations, we fix the stimulation duration to $T=0.4$ seconds as it yields the maximum ITR in both of the datasets. Also, we continue with reporting only the accuracy since it is a direct performance measure with a more intuitive interpretation and the ITR is an invertible function of the accuracy.

\addtocounter{footnote}{-1}
\footnotetext{A large scale version of this figure is given in Supplementary.}

As reported in \autoref{table:sub_eff}, for both of the datasets, using $N_s=3$ sub-bands with $C=9$ channels improves the accuracy by about $2-2.5\%$ compared to using $N_s=1$ sub-band only. Whereas using $2$ more sub-bands with $N_s=5$ neither improves the accuracy nor does it degrade. This indicates that the first three harmonics should be taken into account differently by an appropriate combination as we do in the first layer of the proposed DNN, and harmonics beyond the third degree can be grouped and processed together. Therefore, we continue with using $N_s=3$ sub-bands in the following.

{\autoref{table:ch_eff}  reports the accuracy with $3$ (O1, Oz, O2), $6$ (O1, Oz, O2, POz, PO3, PO4), $9$ (Pz, PO3, PO5, PO4, PO6, POz, O1, Oz, O2) channels as typically used in the literature, and $32$ channels (all channels from occipital, parietal, central-parietal regions and C3, C1, Cz, C2, C4, FCz) as well as all $64$ channels.
Both of using all $64$ channels or $32$ channels improve the accuracy but that also reduce the practicality from the user's point of view. Second, our DNN also works fairly well with $3$ and $6$ channels, which indicates that our algorithm can also be successfully used in the 
more practical systems where only few channels can be used. Overall, using $9$ channels is a reasonable choice in this trade-off between the accuracy and practicality. Hence, we offer the proposed DNN with the settings of $T=0.4$ seconds of stimulation, $N_s=3$ sub-bands and $C=9$ channels.} 

Table \ref{table:req_ch_eff} presents the mean classification accuracy (along with the standard error) achieved by the proposed DNN for stages of our training, in the setting of $T=0.4$ second of stimulation, $N_s=3$ sub-bands and $C=9$ channels. We observe that using only the first stage (our global model) or using only the second stage (directly training our individual models) perform comparably on the BETA dataset but the latter delivers a better performance on the benchmark dataset. On the other hand, using both stages sequentially as proposed by first obtaining a global model and then fine tuning it for each individual model outperforms using only the first stage or using only the second stage on both datasets by $26\%$ on average. This demonstrates the efficacy of our training approach. The same table also records the required running-times per epoch in each case, when run on an NVIDIA GPU (Tesla V100 Volta with memory 32GB). We observe that the second stage training takes negligible time, compared to the first stage taking $\sim 30$ minutes in the case of the benchmark dataset with $1000$ epochs (it takes longer in the case of the BETA dataset). Note that this processing unit is specialized for deep learning algorithms, hence the running times for the first stage would scale up to several hours (completing all $1000$ epochs for a single stimulation duration) on a standard computer of daily use. As for the test time (with $T=0.4$ seconds of stimulation and $C=9$ channels), the classification of a single instance takes about $0.008$ seconds with our DNN whereas, Conv-CA and m-Extended-CCA require about $0.026$ and $0.019$ seconds, respectively, on the same machine.

\textbf{Remark:} In the case of a completely new user, a direct application of our two-stage training strategy requires to (1) first collect a set of training, i.e., calibration, data by conducting new EEG experiments, and then (2) re-train (both two stages of first global DNN model training and fine-tuning it individually afterwards) based on the union of the newly collected data and all the previously existing data. This might be impractical as a limitation as it requires a tedious calibration procedure for a new user. The necessary experiments to collect the calibration data should approximately take only 1-2 hours. However, the first stage of global DNN model training can be lengthy with an unspecialized standard computer since it relies on a large set of data. We point out that the burden of the first stage training can be removed by directly transferring the global DNN model pre-trained on only the existing data (excluding the calibration data of the new user) to the new user, where the calibration data is used only for the fine-tuning of the second stage. Note that the second stage is quite fast since it already starts with a decent initialization and uses only a small amount of data. This alternative training strategy is in fact known as the leave-one-subject-out strategy in the literature, which we propose here as a remedy to that limitation.
\vspace{-3mm}
\subsection{Statistical Significance Analyses}
This part presents our statistical significance test results. Although we achieve a better performance with $64$ channels, the $9$ channels version of our technique is considered in the following for fairness, since all of the other compared methods use $9$ channels. Also, we present ANOVA results where the effect of number of channels and sub-bands are investigated.

For a specific stimulation duration $T$, we conduct $7$ paired t-tests each one of which analyzes the performance difference, observed in Fig. \ref{fig:bench_res} and Fig. \ref{fig:beta_res}, between our proposed DNN ($9$ channels) and one of the $7$ compared algorithms. These tests are repeated for each $T\in \{0.2, 0.4, 0.6, 0.8, 1\}$, and the unadjusted p-values are reported. We call an observed difference ``statistically significant" (*) if the p-value is less than $\frac{0.05}{7}$ (applying ``single" Bonferroni correction by $1/7$ since we have $7$ comparisons for each $T$) and ``statistically highly significant" (**) if the p-value is less than $\frac{0.05}{7\times 5}$ (applying ``double" Bonferroni correction by $1/35$ since we have $35$ comparisons in total across all methods and $T$ choices).

\textbf{In the case of the benchmark dataset:} In terms of the accuracy (Fig. \ref{fig:bench_res}), the least significant difference between our DNN ($9$ channels) and the compared methods is observed with (1) Conv-CA (**$p=3.40 \times 10^{-10}$) for $T=0.2$, (2) eTRCA (**$p=1.04 \times 10^{-6}$) for $T=0.4$, (3) Conv-CA (**$p=5.96 \times 10^{-4}$) for $T=0.6$, (4) eTRCA ($p=0.76 \times 10^{-2}$) for $T=0.8$, and (5) Conv-CA ($p=6.81 \times 10^{-2}$) for $T=1$. Here, for $T=0.8$, the difference with eTRCA is not significant; but it is significant (*) with Conv-CA and ms-eTRCA, and highly significant (**) with all the others. For $T=1$, the difference with Conv-CA, ms-eTRCA and eTRCA are not significant; but it is significant (*) with TSCORRCA, and highly significant (**) with all the others. In terms of ITR (Fig. \ref{fig:bench_res}), the least significant difference between our DNN ($9$ channels) and the compared methods is observed with (1) Conv-CA (**$p=7.38 \times 10^{-10}$) for $T=0.2$, (2) eTRCA (**$p=3.60 \times 10^{-7}$) for $T=0.4$, (3) Conv-CA (**$p=2.55 \times 10^{-4}$) for $T=0.6$, (4) eTRCA (*$p=0.27 \times 10^{-2}$) for $T=0.8$, and (5) eTRCA ($p=4.94 \times 10^{-2}$) for $T=1$. Here, for $T=0.8$, the difference is significant (*) with ms-eTRCA and eTRCA, and highly significant (**) with all the others. For $T=1$, the difference with Conv-CA, ms-eTRCA and eTRCA are not significant; but it is significant (*) with TSCORRCA, and highly significant (**) with all the others.

\textbf{In the case of the BETA dataset:} {In terms of the accuracy (Fig. \ref{fig:beta_res}), the least significant difference between our DNN ($9$ channels) and the compared methods is observed with (1) Conv-CA (**$p=5.13 \times 10^{-13}$) for $T=0.2$, (2) Conv-CA (**$p=3.58 \times 10^{-13}$) for $T=0.4$, (3) Conv-CA (**$p=3.49 \times 10^{-10}$) for $T=0.6$, (4) Conv-CA (**$p=3.80 \times 10^{-9}$) for $T=0.8$, and (5) Conv-CA ($**p=3.64 \times 10^{-7}$) for $T=1$. In terms of ITR (Fig. \ref{fig:beta_res}), the least significant difference between our DNN ($9$ channels) and the compared methods is observed with (1) Conv-CA (**$p=3.88 \times 10^{-12}$) for $T=0.2$, (2) Conv-CA (**$p=1.50 \times 10^{-13}$) for $T=0.4$, (3) Conv-CA (**$p=1.75 \times 10^{-10}$) for $T=0.6$, (4) Conv-CA (**$p=3.84 \times 10^{-10}$) for $T=0.8$, and (5) Conv-CA (**$p=1.32 \times 10^{-7}$) for $T=1$. Thus, the difference (in terms of both accuracy and ITR) is always highly significant (**) with all the other compared methods and for all $T$'s.}

On the other hand, a one-way repeated measures ANOVA reveals a main effect of the number of sub-bands on the accuracy (Benchmark: $F(4,136)=30.271$, $p<5.18\times 10^{-18}$; BETA: $F(4,276)=39.793$, $p<2.64\times 10^{-26}$) with our proposed DNN in Table \ref{table:sub_eff}, where a paired t-test indicates a highly significant difference between using $1$ sub-band and $3$ sub-bands (Benchmark: $p=3.05\times 10^{-9}$; BETA: $p=6.55\times 10^{-13}$). Similarly, the number of channels (Table \ref{table:ch_eff}) has a main effect on the accuracy (Benchmark: $F(4,136)=86.007$, $p<2.82\times 10^{-36}$; BETA: $F(4,276)=162.24$, $p<3.23\times 10^{-71}$), where a paired t-test indicates a highly significant difference between using $9$ channels and $64$ channels (Benchmark: $p=5.29\times 10^{-4}$; BETA: $p=3.89\times 10^{-4}$). The training strategy (Table \ref{table:req_ch_eff}) also has a main effect (Benchmark: $F(2,68)=87.466$, $p<1.58\times 10^{-19}$; BETA: $F(2,138)=218.08$, $p<1.901\times 10^{-43}$), where a paired t-test indicates a highly significant difference between employing only the second stage and two-stage (Benchmark: $p=1.13\times 10^{-13}$; BETA: $p=2.93\times 10^{-37}$).
\newcolumntype{P}[1]{>{\centering\arraybackslash}p{#1}}
\begin{table}[t!]
\centering
\caption{The mean classification accuracy ($\%$), with the standard error, of our DNN is reported versus varying number of sub-bands with $9$ channels and $0.4$ seconds of stimulation.}
\begin{tabular}{m{1.8cm}|P{2.5cm}|P{2.5cm} |} 
\cline{2-3}
&Benchmark \cite{wang2016benchmark}& BETA \cite{BETA-data} \\ \hline 
\multicolumn{1}{|l|}{1 sub-band} & $77.89\pm2.89$ &  $64.98\pm2.25$\\ \hline 
\multicolumn{1}{|l|}{2 sub-bands} & $78.80\pm2.92$ &  $66.84\pm2.17$\\ \hline
\multicolumn{1}{|l|}{3 sub-bands} & $79.89\pm2.81$ &  $67.52\pm2.17$\\ \hline
\multicolumn{1}{|l|}{4 sub-bands} & $79.86\pm2.89$ &  $67.54\pm2.12$\\ \hline
\multicolumn{1}{|l|}{5 sub-bands} & $79.96\pm2.82$ &  $67.66\pm2.17$\\ \hline 
\end{tabular}
\label{table:sub_eff}
\vspace{-3mm}
\end{table}
\begin{table}[t!]
\centering
\caption{The mean classification accuracy ($\%$), with the standard error, of our DNN is reported versus varying number of channels with $3$ sub-bands and $0.4$ seconds of stimulation.}
\begin{tabular}{m{1.8cm}|P{2.5cm}|P{2.5cm} |} 
\cline{2-3}
&Benchmark \cite{wang2016benchmark}& BETA \cite{BETA-data} \\ \hline 
\multicolumn{1}{|l|}{3 channels} & $51.04\pm4.09$ &  $42.73\pm2.60$\\ \hline 
\multicolumn{1}{|l|}{6 channels} & $74.55\pm3.12$ &  $58.86\pm2.49$\\ \hline
\multicolumn{1}{|l|}{9 channels} & $79.89\pm2.81$ &  $67.52\pm2.17$\\ \hline
\multicolumn{1}{|l|}{32 channels} & $82.70\pm2.63$ &  $70.21\pm2.05$\\ \hline
\multicolumn{1}{|l|}{64 channels} & $84.54\pm2.08$ &  $69.54\pm2.07$\\ \hline
\end{tabular}
\label{table:ch_eff}
\vspace{-3mm}
\end{table} 
\begin{table}[t!]
\centering
\caption{The mean classification accuracy ($\%$) of our DNN is reported along with the standard error for the stages of our training. Running-times per epoch are also provided below each line.}
\begin{tabular}{m{1.8cm}|P{2.5cm}|P{2.5cm} |} 
\cline{2-3}
&Benchmark \cite{wang2016benchmark}& BETA \cite{BETA-data} \\ \hline 
\multicolumn{1}{|l|}{Only first stage} & $46.93\pm3.30$ &  $38.44\pm2.12$\\ 
\multicolumn{1}{|l|}{} & ($\sim 1.80$ sec.) &  ($\sim 2.25$ sec.)\\ \hline 
\multicolumn{1}{|l|}{Only second stage} & $56.39\pm4.08$ &  $34.45\pm2.43$\\ 
\multicolumn{1}{|l|}{} & ($\sim 0.055$ sec.) &  ($\sim 0.05$ sec.)\\\hline
\multicolumn{1}{|l|}{Two-stage} & $79.89\pm2.81$ &  $67.52\pm2.17$\\ 
\multicolumn{1}{|l|}{} & ($\sim 1.855$ sec.) &  ($\sim 2.30$ sec.)\\\hline
\end{tabular}
\label{table:req_ch_eff}
\vspace{-3mm}
\end{table}

\vspace{-3mm}
\subsection{Error Patterns}
\begin{figure}[t]
\includegraphics[width=0.47\textwidth]{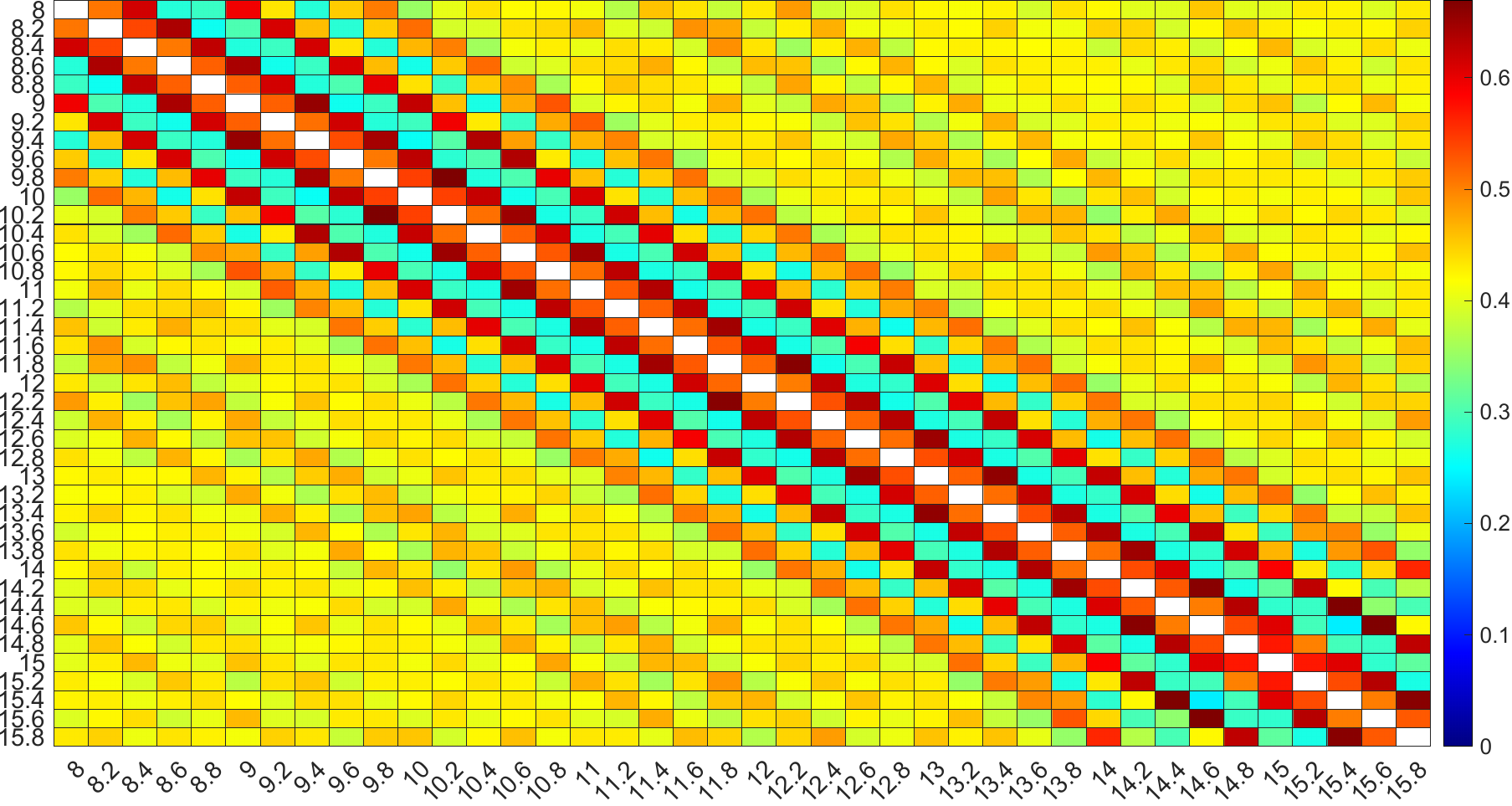}
\centering
\caption{The matrix $M_D$ of the mean absolute distances for any pair of contrast modulating sinusoids with frequencies $\{8,8.2,\cdots,15.8\}$ that are used for frequency tagging in the BCI SSVEP spellers of the both datasets \cite{wang2016benchmark,BETA-data}. The distance is the smallest when two frequencies are $0.6$ or $0.8$ Hz apart, and the largest when $0.2, 0.4$ or $1$ Hz apart.\textsuperscript{2}}
\label{fig:sin_dist}
\vspace{-4mm}
\end{figure}

Considering the inter-class confusions presented in \autoref{fig:bench_conf} and \autoref{fig:beta_conf}, we observe two prominent error patterns. The first pattern is that there exists a pronounced rate of error diagonally along the two lines $f_{\text{true}}=f_{\text{predict}}\pm 0.6$, but this pattern completely disappears at even the adjacent closest neighbors of $f_{\text{true}}=f_{\text{predict}}\pm 0.2$ or $f_{\text{true}}=f_{\text{predict}}\pm 0.4$. This is surprising as one would expect to confuse the target character with the character of the closest frequency. The reason we discover for this finding is the following.  Firstly, we define the mean absolute distance $M_D(i,j)$ between two contrast modulating sinusoids with frequencies $f_i$, $f_j$ from the set $\{8,8.2,\cdots,15.8\}$ and  phases $\theta_i$, $\theta_j$ from the set $\{0, 0.5\pi, \pi, 1.5\pi\}$ as the following. 
\begin{align}
     M_D(i,j) &= \frac{1}{T \times R}\sum_{k=0}^{T \times R - 1} |s(f_i,\theta_i,k)-s(f_j,\theta_j,k)|, 
\end{align}
where $R=60$ Hz is the refresh rate of the monitor, $k$ is the discrete time index, $T$ is the stimulation duration, and the contrast modulating sinusoid is defined as $s(f,\theta,k)= \frac{1}{2}(1+\sin(2\pi f k/R+\theta))$ (see the dataset descriptions in \cite{wang2016benchmark} or \cite{BETA-data}). The distance between the samples, which multiply the luminance of the character thumbnails as shown in \autoref{fig:bcisystem}, from two contrast modulating sinusoids with the frequencies $f_1$ and $f_2$ is the smallest when the frequencies are chosen as $f_2=f_1\pm 0.6$ whereas it is the largest when chosen as $f_2=f_1\pm 0.2$ or $f_1\pm 0.4$. This is demonstrated in the matrix of distances in \autoref{fig:sin_dist}, and please see the strong correlation between the pattern in the \autoref{fig:sin_dist} and the pattern in the confusion matrices of \autoref{fig:bench_conf} and \autoref{fig:beta_conf}.  Consequently, during stimulation, the luminance variations falling onto the retina and the corresponding early projections to the visual cortex are maximally similar when the frequencies are $f_1$ and $f_1\pm0.6$ and maximally dissimilar when the frequencies are $f_1$ and $f_1\pm0.2$ or $f_1\pm0.4$. This similarity appears to reduce the discrimination power, hence negatively affects the performance.       

This first pattern is perhaps best understood with the vertical intermodulations (IM) resulting from the layout used in the character matrix, which emerges in our study as the second error pattern that is uniformly observed in the confusion matrix (\autoref{fig:bench_conf}) of the benchmark dataset. The source of confusion by this second error pattern is the well-known IM phenomenon (cf. \cite{alp2018measuring} and the references therein) which generates the IM components in the SSVEP spectrum at the integer multiples of the spatially nearby flickering frequencies that the subject is exposed to. If an IM is generated that is close to the frequency of a non-target character and also close to the one of the target character itself, then there seems to happen an error due to confusion. Since this is consistently and strongly observed in the confusion matrix (\autoref{fig:bench_conf}) of the benchmark dataset, we present as an important finding of our study that the vertical IM, together with the first pattern of sinusoidal distances, has an important effect on how the errors happen in an emphasizing-the-first-pattern (if the IM exists) or suppressing-the-first-pattern (if the IM does not exist) manner. Namely, the relatively large rate of confusion in the first pattern between the frequencies $f_1$ (predicted frequency) and $f_2$ (true frequency) is persistent, A) only when the target character ($f_2$ for $f_2=f_1+0.6$) is on the first, second or the third rows of the character matrix (\autoref{fig:bench_screen}) and also B) only when the target character ($f_2$ for $f_2=f_1-0.6$) is on the third, fourth or fifth rows. We attribute this to the interference by the fifth degree vertical IM generated in both the cases A and B: In the case A, the confusing predicted frequency $f_1$ can be obtained as the $5$th IM $f_1=3\times f_2 -f_2^{l1} -f_2^{l2}$, and in the case B, $f_1$ can be obtained as the $5$'th IM $f_1=3\times f_2 -f_2^{u1} -f_2^{u2}$.  Here, $l_k$ or $u_k$ are the $k^{th}$ lower/upper adjacent frequency in the character matrix for $k=1$ or $k=2$. For example, the true (target) frequency $f_2=14.2$ Hz (character ``O" on the second row in Fig. \ref{fig:bcisystem} and Fig. \ref{fig:bench_screen}) has the two lower neighbors $f_2^{l1}=14.4$ Hz and $f_2^{l2}=14.6$ Hz, generating the $5$'th degree IM of case A as $3\times f_2 -f_2^{l1} -f_2^{l2}=3\times 14.2 -14.4 -14.6=13.6$ Hz. Since this IM component appears in the received EEG signal and when this existence is strong enough, it is predicted as the true frequency, i.e., $f_1=13.6$ Hz (character ``3"), which actually corresponds to the most frequent error ($16$ times of confusion $14.2$ Hz vs $13.6$ Hz) in Fig. \ref{fig:bench_conf}. Such errors can be visually traced by noting the colored patterns below (case A) and above (case B) the diagonal of the confusion matrix in Fig. \ref{fig:bench_conf}.  Note that we do not observe a detrimental IM when the target character is on the first or second rows in the case A and if it is on the fourth and fifth rows in the case B, namely, if it does not have two vertical adjacent neighbors.  Also, a detrimental IM is not observed horizontally or also not observed at lower degrees (lower than $5$) because of not that the horizontal and lower degree IMs are not generated (they are certainly generated) but that they are not any close to the frequency of the target character. Hence, an error does not happen in that specific way. We emphasize that the intermodulation effect is specific to the character matrix used in the stimulation. For this reason, we do not observe the same exact second pattern in the keyboard presentation of the BETA dataset (cf. \autoref{fig:beta_conf}) as the characters are shuffled on the matrix. \textit{These two error patterns can provide key insights to the matrix and stimulation design in future studies.} 


\begin{figure}[t!]%
\centering
\includegraphics[width=0.47\textwidth]{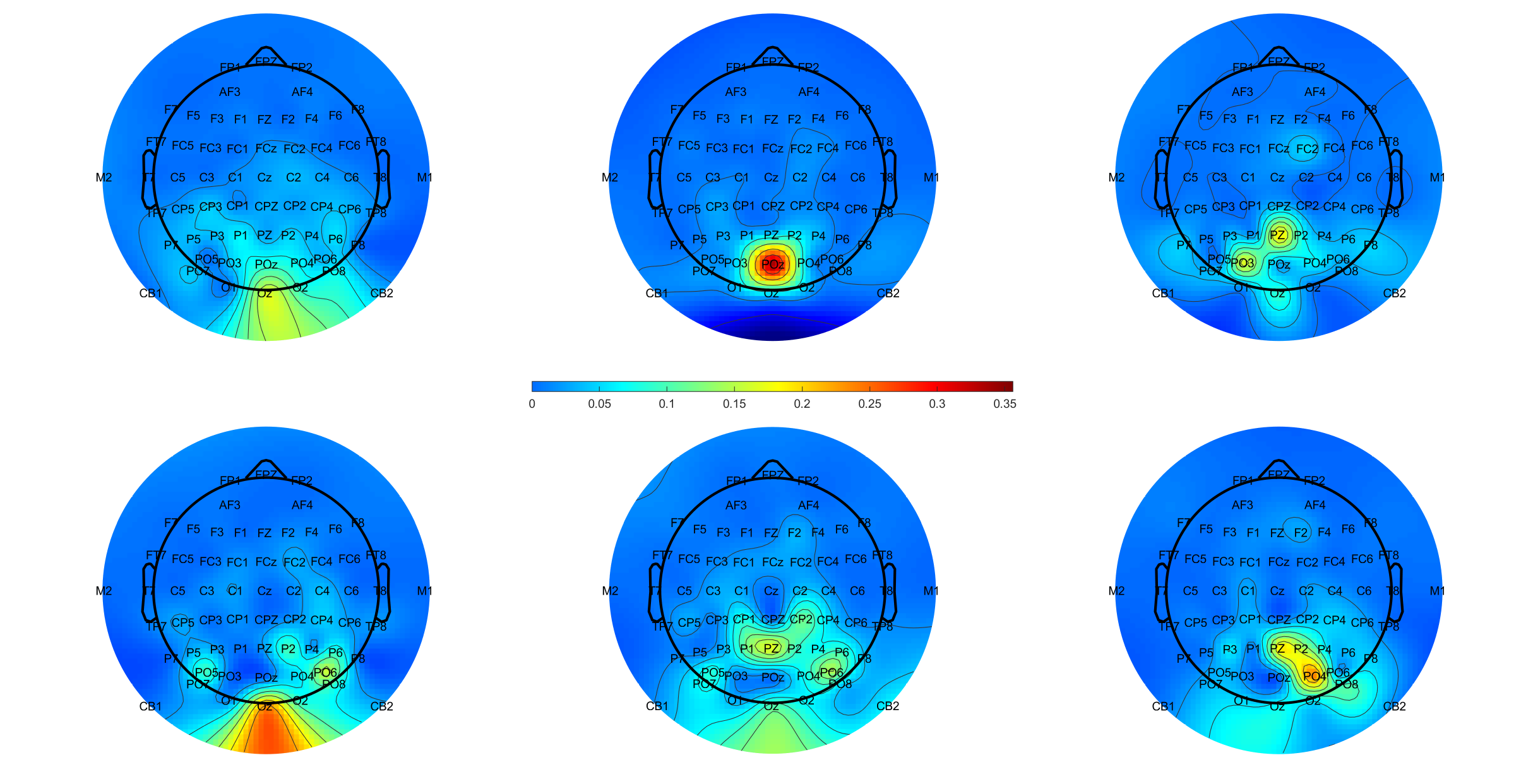}
\caption{ Upper row (and the lower row) presents the topographic map of the $3$ channel combinations, in no particular order, learned by the proposed DNN in the case of the benchmark (and the BETA) dataset.\textsuperscript{2} The maps are generated by EEGLAB \cite{eeglab}.} 
\label{fig:topoplot}
\vspace{-3mm}
\end{figure}

\vspace{-2mm}
\subsection{Topographic Maps}
Lastly, we study the importance of the electrodes (i.e. channels), in terms of their contribution to the target identification accuracy, by analyzing the channel combinations learned by the proposed DNN. For this purpose, we concentrate on only $3$ combinations (since analyzing all $120$ combinations given by the network would not be practical) such that for instance a large weight (in absolute value) in the combinations for a channel indicates a high importance. In the current setting, our network learns $120$ channel combinations in its second layer (cf. \autoref{fig:bcisystem}) for the best achievable accuracy but does not rank them with respect to their importance to the accuracy. Hence, currently, it is not possible (without a post-component analysis of the $120$ combinations which is not in the scope of the presented work) to immediately choose the most important $3$ combinations out of those $120$ ones. Nevertheless, in this part only, in order to have the network determine the $3$ combinations, we set the size of the second layer output as $1\times 50 \times 3$ instead of the original size, and then train the network based on the entire set of available data when fed with all available $64$ channels. Based on this approach, \autoref{fig:topoplot} presents the topographic maps of the $3$ channel combinations (without ranking them) that are learned by the proposed DNN in the both datasets. In \autoref{fig:topoplot}, each channel has a color closer to red (blue) if it has proportionally higher (lower) weight in the absolute value. We first observe that the channels that are recommended in the study \cite{deep2} are also mostly covered by our network, which is an independent verification of a previous result. Therefore, this indicates that our network does not or limitedly overfit. Additionally, the important channels are more concentrated in the case of the benchmark dataset, whereas they are spread more in the case of the BETA dataset. We attribute this to the low SNR of the BETA dataset. Therefore, unlike \cite{deep2}, when the SNR is low, we tend to recommend more channels from the parietal region (such as P1, P2). Whereas we strongly recommend the channels Oz and Pz as they are shared by the both datasets. Also, our topographic maps are complementary to each other indicating the necessity of combining the channels, and that is to be nonlinearly because a totally linear approach could exploit combinations as many as the number of channels.

\vspace{-1.5mm}
\section{Conclusion} \label{sec:CO}
{ We study the target identification of BCI SSVEP spellers, which is a multi-class classification problem with $40$ classes where the goal is to classify the SSVEP signal received through EEG during an experiment, thereby predicting the target character that subject intends to spell. To this end, we proposed a novel DNN architecture that consists of $4$ convolutional (sub-band and channel combinations as well as downsampling and filtering in time) and $1$ fully connected layers. The proposed DNN strongly outperforms the state-of-the-art as well as the most recently proposed techniques in the literature on two publicly available large scale benchmark and BETA datasets. We achieve ITRs with only $0.4$ seconds of stimulation: \textit{$265.23$ bits/min} on the benchmark and \textit{$196.59$ bits/min} on the BETA dataset. To our best knowledge, these are the highest (and significantly larger than the nearest competitor) performance results ever reported on these datasets.
\vspace{-4mm}
}

%




%
\bibliographystyle{IEEEtran}

%



\enlargethispage{-5in}

\end{document}